\def\papertitle{Diet deep generative audio models with structured lottery}
\def\paperauthorA{Philippe Esling}
\def\paperauthorB{Ninon Devis}
\def\paperauthorC{Adrien Bitton}
\def\paperauthorD{Antoine Caillon}
\def\paperauthorE{Axel Chemla--Romeu-Santos}
\def\paperauthorF{Constance Douwes}

\documentclass[twoside,a4paper]{article}
\usepackage{etoolbox}
\usepackage{dafx_20}
\usepackage{amsmath,amssymb,amsfonts,amsthm}
\usepackage{euscript}
\usepackage[T1]{fontenc}
\usepackage[utf8]{inputenc}
\usepackage{nimbusserif}
\usepackage{ifpdf}
\usepackage[english]{babel}
\usepackage{caption}
\usepackage{subfig} 
\usepackage{color}
\usepackage{maths}

\input glyphtounicode
\ninept

\newcounter{numauth}
\newcounter{listcnt}
\newcommand\authcnt[1]{\ifdefined#1 \stepcounter{numauth} \fi}

\newcommand\addauth[1]{
\ifdefined#1 
\stepcounter{listcnt}
\ifnum \value{listcnt}<\value{numauth}
\appto\authorslist{, #1}
\else
\appto\authorslist{~and~#1}
\fi
\fi}
\authcnt{\paperauthorB}
\authcnt{\paperauthorC}
\authcnt{\paperauthorD}
\authcnt{\paperauthorE}
\authcnt{\paperauthorF}
\authcnt{\paperauthorJ}
\def\authorslist{\paperauthorA}
\addauth{\paperauthorB}
\addauth{\paperauthorC}
\addauth{\paperauthorD}
\addauth{\paperauthorE}
\addauth{\paperauthorF}

\usepackage{times}

\newif\ifpdf
\ifx\pdfoutput\relax
\else
   \ifcase\pdfoutput
      \pdffalse
   \else
      \pdftrue
\fi

\usepackage[hypcap=true]{caption}
\title{\papertitle}

\affiliation{
\paperauthorA, \paperauthorB, \paperauthorC, \paperauthorD, \paperauthorE, \paperauthorF \,\sthanks{This work is supported by the ANR:17-CE38-0015-01 MAKIMOno project, the SSHRC:895-2018-1023 ACTOR Partnership and Emergence(s) ACIDITEAM project from Paris and ACIMO from Sorbonne.\vspace{1mm}}}
{\href{https://www.ircam.fr/}{Institut de Recherche et Coordination Acoustique / Musique (IRCAM) - Sorbonne Université, CNRS UMR 9912} \\ 1, place Igor Stravinsky, Paris, France\\
{\tt \href{mailto:esling@ircam.fr}{esling@ircam.fr}}
}

\begin{document}
\ifpdf 
  \DeclareGraphicsExtensions{.png,.jpg,.pdf}
\else  
  \DeclareGraphicsExtensions{.eps}
\fi


\maketitle

\begin{abstract}
Deep learning models have provided extremely successful solutions in most audio application fields. However, the high accuracy of these models comes at the expense of a tremendous computation cost. This aspect is almost always overlooked in evaluating the quality of proposed models. However, models should not be evaluated without taking into account their complexity. This aspect is especially critical in audio applications, which heavily relies on specialized embedded hardware with real-time constraints.

In this paper, we build on recent observations that deep models are highly overparameterized, by studying the \textit{lottery ticket hypothesis} on deep generative audio models. This hypothesis states that extremely efficient small sub-networks exist in deep models and would provide higher accuracy than larger models if trained in isolation. However, lottery tickets are found by relying on unstructured \textit{masking}, which means that resulting models do not provide any gain in either disk size or inference time. Instead, we develop here a method aimed at performing \textit{structured trimming}. We show that this requires to rely on \textit{global} selection and introduce a specific criterion based on mutual information.

First, we confirm the surprising result that \textit{smaller models} provide \textit{higher accuracy} than their large counterparts. We further show that we can remove up to $95\%$ of the model weights without significant degradation in accuracy. Hence, we can obtain very light models for generative audio across popular methods such as \textit{Wavenet}, \textit{SING} or \textit{DDSP}, that are up to 100 times smaller with commensurate accuracy. We study the theoretical bounds for embedding these models on Raspberry Pi and Arduino, and show that we can obtain generative models on CPU with equivalent quality as large GPU models. Finally, we discuss the possibility of implementing deep generative audio models on embedded platforms.\footnote{All results, code and embedded library are available on the supporting webpage to this article \href{https://github.com/acids-ircam/lottery_generative}{https://github.com/acids-ircam/lottery\_generative}.\vspace{1mm}}

\end{abstract}

\section{Introduction}
\label{sec:intro}
Over the past years, deep learning models have witnessed tremendous success in a wide variety of applications.  Specifically, in the audio signal domain, novel \textit{deep generative models} \cite{mehri2016samplernn} are able to synthesize waveform data matching the acoustic properties of a given dataset with unprecedented quality. This specific task is highly challenging as the generation of high-quality waveform requires to handle complex temporal structures at both local and global scales. Nevertheless, recent auto-regressive (WaveNet \cite{van2016wavenet}) or recurrent (SampleRNN \cite{mehri2016samplernn}) architectures successfully model raw audio waveform. In order to attain this goal, all approaches require extremely complex architectures with humongous numbers of parameters. This incurs significant energy and computational costs along with huge memory footprints. Unfortunately, the complexity of resulting solutions and their extensive inference time are almost systematically overlooked properties, obliviated by the never-ending quest for accuracy. However, this goal becomes paramount when aiming to provide these systems to users in real-time settings or on dedicated lightweight embedded hardware, which are particularly pervasive in the audio generation domain. Subsequently, none of the current deep generative audio models can fit these computational constraints or memory limitations. 

In parallel, it has been repeatedly observed that deep architectures are profoundly over-parameterized. This implies that a large majority of the parameters in deep models could potentially be removed without significant loss in performance \cite{belkin2019reconciling}. However, this over-parameterization appears to be required for correctly training deep models, as it allows the optimization process to search for solutions in a simpler landscape \cite{arora2018optimization}. The idea of \textit{pruning} \cite{lecun1990optimal} has been proposed to remove the less relevant weights from a trained model in order to reduce its size. Unfortunately, the pruning approach usually only provides small compression ratios, in order to avoid large losses in accuracy \cite{liu2018rethinking}. The recently proposed \textit{lottery ticket hypothesis} \cite{frankle2019lottery} conjectures the existence of extremely efficient sparse sub-networks within very large models, already existing at initialization. These sub-networks could be able to reach a similar, or even higher, accuracy if they were trained in isolation and their weights are \textit{rewound} to earlier epochs of training \cite{frankle2020the}. Furthermore, it appears possible to mask up to 99.5\% of the model weights without significant loss in accuracy, providing extremely sparse solutions to the same task. Unfortunately, this approach relies on \textit{masking} selected weights (called \textit{unstructured pruning}), thus maintaining both the size and inference costs of large models. 

In this paper, we propose to build on the lottery ticket hypothesis by extending its use to \textit{structured} scenarios. In these, we seek to remove entire units of computation (equivalently convolutional \textit{channels}), instead of only masking the corresponding weights. This would allow to truly reduce the model size and correspondingly its inference time. Hence, we first perform an evaluation of the original lottery ticket framework for generative audio models, by implementing and pruning several state-of-art deep generative audio models, such as \textit{Wavenet} \cite{van2016wavenet}, \textit{SING} \cite{defossez2018sing} and \textit{DDSP} \cite{engel2020ddsp}. We show that the original lottery results hold for generative models, but in a lower extent than discriminative tasks. Still, we confirm the surprising results that we obtain higher accuracy by masking up to $60\%$ of the original weights, and we can maintain the original accuracy, even when masking up to $95\%$ of the weights. Based on this, we show that even though we are able to mask a stunningly large portion of the network, we can effectively remove only a small portion of the computation units. To alleviate this issue, we introduce several pruning criteria that can account for \textit{global} pruning scenarios. Indeed, we hypothesize that performing \textit{local structured} pruning only amounts to an expensive form of architecture search (as we reduce all layers in the network by an identical amount). Oppositely, performing \textit{global structured} pruning could allow to leverage the creation of bottleneck layers along the architecture. In order to take full advantage of this idea, we propose a specific criterion based on \textit{information-theoretic} principles. We show that computational units that \textit{globally maximize} the mutual information with respect to the target are able to withstand a large level of compression, while maintaining high accuracy. We evaluate our proposal on several audio generative models, by assessing their memory, size and inference time (FLOPs) requirements. We show that we can obtain ultra-light generative audio models able to perform real-time inference on CPU, with quality comparable to very large GPU models. Finally, we define and study the \textit{real-time} and \textit{embeddable} bounds of our ultra-light generative audio models, in order to fit the requirements of the Raspberry and Arduino platforms. We show that deep models can be embeddable on Raspberry and discuss a library for performing embedded deep audio generation.

\section{State-of-art}
\label{sec:stateofart}


\subsection{Audio waveform generation} \label{sec:audio_soa}

In order to leverage deep neural networks for audio synthesis, several approaches have first targeted the generation of spectral information, by relying on either variational auto-encoders \cite{esling2018generative} or generative adversarial networks \cite{engel2019gansynth}. However, spectral representations suffer from multiple drawbacks in generative setups. Notably, learning schemes preclude the use of phase information, which forces to rely on approximate phase reconstruction algorithms \cite{perraudin2013fast}, degrading the generation quality.

To address these limitations, several models have directly targeted raw audio waveform, which must face the high sampling rates and temporal complexity of such data. Indeed, these models must process simultaneously local features to ensure audio quality, while being able to analyze longer-term dependencies in order to generate coherent audio information. Given an audio waveform $\bx=\{x_1,\ldots,x_T\}$, we can define the joint distribution $p(\bx)$ as a product of conditional distributions, through the causality assumption that each sample is only dependent on the previous ones
\begin{equation}
p(\mathbf{x})=\prod_{t=1}^{T}p(x_t|x_1,\ldots,x_{t-1}).
\label{eq:wavenet}
\end{equation}
Following this auto-regressive formulation, \textit{Wavenet} \cite{van2016wavenet} casts the problem of predicting waveform samples as a classification task over amplitudes with a $\mu$-law quantization. This model is able to handle long-term dependencies by using stacked layers of gated \textit{dilated} convolutions, which exponentially increase the receptive field of the model. This approach is now an established solution for high-quality speech synthesis and has also been successfully applied to musical audio with the NSynth dataset \cite{engel2017nsynth}. However, auto-regressive modeling is inherently slow since the samples can only be processed iteratively. Moreover, large convolution kernels and numbers of layers are required to infer even small-sized contexts. This results in computationally heavy models, with large inference and training times. Based on similar assumptions, \textit{SampleRNN} \cite{mehri2016samplernn} relies on recurrent networks, performing computation in a hierarchical manner. Multiple temporal scales are defined through a sample-level auto-regressive module and a longer-term network. Although this model provides convincing audio results, it still incurs a heavy computational cost.

More recent streams of research rely on the differentiability of the STFT to compute losses in the spectral domain, rather than directly on waveform samples. This allows to produce different waveforms with equivalent spectral content at a low computational cost. Given a signal $\bx$ with spectrogram $S_{w}(\bx)=\left|\text{STFT}_{w}[\mathbf{x}]\right|^{2}$, computed on a window $w$, the multiscale learning loss is
\begin{equation}
\underset{\btheta}{\text{argmin }}\sum_{i}\norm{\log\left(S_{w_{i}}(\mathbf{x}) + \bepsilon \right),\log\left(S_{w_{i}}(\hat{\mathbf{x}}) + \bepsilon \right)}_{1}
\label{eq:SING}
\end{equation}
where $\hat{\mathbf{x}}$ is the waveform produced by the model with parameters $\btheta$. Based on this idea, the Symbol-to-Instrument Neural Generator (SING) \cite{defossez2018sing} relies on an overlap-add convolutional architecture, which constructs sequences of overlapping audio segments. The model processes signal windows to reduce the input dimensionality entering an up-sampling convolutional decoder. A top-level sequential embedding is trained on frames, by conditioning over instrument, pitch and velocity classes. Given this specific architecture, the model is highly constrained to produce only individual pitched instrumental notes of fixed duration. Several models have extended this idea, by relying on stronger assumptions and inductive biases from digital signal processing. First, the Neural Source-Filter (NSF) model \cite{wang2019neural} splits the generation between successive source and filtering modules, mimicking traditional source-filter models. Hence, a sinusoidal (voiced) and noise (unvoiced) excitations are fed into separate filter modules, allowing to model different types of signals. Similar to NSF, the Differentiable Digital Signal Processing (DDSP) model \cite{engel2020ddsp} has been proposed to target pitched musical audio. This architecture similarly implements an harmonic additive synthesizer summed with a filtered noise synthesizer (defined as a trainable Finite Impulse Response filter). This approach can be seen as a form of amortization, that learns to control a synthesis process based on fundamental frequency, loudness and latent features extracted from the input waveform. 

Despite the successes provided by these models, they still require large computational costs, only handled by modern GPUs. Furthermore, these also entail large disk and memory usage, precluding any use of these models on embedded devices. This heavily limits the use of deep networks in audio applications, which are mostly real-time, on specific lightweight and non-GPU hardware.

\subsection{Overparameterization of learning models} \label{sec:overparameterization}

\subsubsection{Model compression and pruning}
The idea of \textit{model compression} in neural networks has been proposed quite early, with the pioneering approach of \textit{pruning} proposed by LeCun \cite{lecun1990optimal}. The underlying idea is to remove the weights that least contribute to the accuracy of the network, as defined by a pruning criterion. This method, which is still amongst the most widespread, is based on three steps: (i) \textit{training} a large over-parameterized network, (ii) \textit{removing} weights according to a given criterion and (iii) \textit{fine-tuning} the remaining weights to restore the lost performance. Several methods have since been proposed and can be broadly divided between \textit{structured} and \textit{unstructured} pruning criterion. \textit{Structured} pruning aims to remove structural parts of a network (such as entire convolutional channels), whereas \textit{unstructured} pruning acts directly on individual parameters, regardless of the underlying architecture.

\textbf{Structured Pruning}. Approaches in structured pruning mostly revolve around the idea of removing \textit{channels} in convolutional layers. With that aim, several criteria for determining the channels to remove were proposed, such as computing the $L_{n}$-norm of different filters \cite{li2016pruning}. Although structured pruning can allow to remove large parts of a network, it remains at largely lower compression and accuracy than unstructured methods \cite{liu2018rethinking}. 

\textbf{Unstructured Pruning}. Most of the proposed pruning methods are based on the magnitude of individual parameters \cite{lecun1990optimal}, even in the case of convolutional networks \cite{han2015learning}. In these, the pruned weights are \textit{masked} instead of being removed, leading to sparse weight matrices but with identical dimensionality. The advantage of this masking approach is that it allows to handle any type of layer indistinctly. However, the resulting model does not provide any gain in size or inference time.

Finally, it should be noted that most pruning methods require multiple trainings. In order to save training costs, some methods aim to gradually prune the model across a single training phase \cite{lee2018snip}. However, these approaches appear to be less efficient than their multiple training counterparts \cite{frankle2020the}.



\subsubsection{Lottery ticket hypothesis} \label{sec:lottery}

The \textit{lottery ticket hypothesis}~\cite{frankle2018lottery} conjectures the existence of extremely efficient sparse sub-networks already present in randomly initialised neural networks.  Those sub-networks, called \textit{winning tickets} (WT), would provide higher accuracy than their large counterparts if they were trained in isolation, while allowing for massive amounts of pruning. Those WT are based on initial weights and connectivity patterns with "lucky initialisation" that lead to particularly effective training. Identification of the WT is performed by first fully training the network and, then, masking the smallest-magnitude weights. The structure of the WT is defined by the unpruned weights, which are subsequently \textit{reset to their initialisation values} and retrained. This procedure is repeatedly applied, leading to Iterative Magnitude Pruning (IMP). On the MNIST and CIFAR sets, removing up to 80\% of the weights provide higher accuracy than the original network, while the original accuracy can be maintained even when removing up to 96.5\% of the weights. 

The \textit{reset} operation is a crucial step in IMP as randomly re-initialised tickets were shown to reach lower accuracy than the original large network. In a further study for deeper networks \cite{frankle2019lottery}, the authors showed that it was beneficial to \textit{rewind} the weights at a given early epoch in the training, rather than at initialization values. Lottery tickets could still be uncovered in deeper architectures only by relying on this \textit{rewinding} operation. 



Formally, a network is defined as a function $f(\mathbf{x};\mathbf{W})$ with weights $\mathbf{W}$ randomly initialized as $\mathbf{W}_0 \sim p(\mathbf{W})$. The network is trained to reach accuracy $a^*$ in $T^*$ iterations with final weights $\mathbf{W}_{T^*}$. A \textit{sub-network} can be seen as a tuple $(\mathbf{W}, \mathbf{M})$ of weight values $\mathbf{W} \in R^D$ and a pruning mask $\mathbf{M}\in\{0,1\}^{|\mathbf{W}|}$, defining the function $f(\mathbf{x};\mathbf{M}\odot\mathbf{W})$, where $\odot$ is the element-wise product. The \textit{lottery ticket hypothesis} states that there exists a sub-network $(\mathbf{W}_k, \mathbf{M})$ with a given pruning mask $\mathbf{M}$, and iteration $k \ll T^*$, such that retraining this sub-network will reach a \textit{commensurate accuracy} $a \geq a*$ in \textit{commensurate training time} $T \leq T^*-k$, with \textit{fewer parameters} $||\mathbf{M}|| \ll |\mathbf{W}|$. Given these definitions, IMP training with rewinding can be implemented as follows


\begin{enumerate}
    \item \textit{Initialisation}. Initialise parameters $\mathbf{W}_0$ randomly and the mask \textbf{M} to all ones, defining the network $f(\mathbf{x};\mathbf{M}\odot\mathbf{W}_{0})$.
    \item \textit{Network training}. Train the parameters $\mathbf{W}_{i}$ of the network $f(\mathbf{x};\mathbf{M}\odot\mathbf{W}_{i})$ to completion $\mathbf{W}_{T^*}$, while storing the weights $\mathbf{W}_k$ at iteration $k$, by performing
    \begin{enumerate}
        \item Train the weights for $k$ iterations, producing the network $f(\mathbf{x};\mathbf{M}\odot\mathbf{W}_k)$.
        \item Train the network for $T^{*}-k$ further iterations, producing the network $f(\mathbf{x};\mathbf{M}\odot\mathbf{W}_{T^{*}})$.
    \end{enumerate}
    \item \textit{Weight Selection}. Prune the weights $\mathbf{W}_{T^*}$ by computing a masking criterion $\mathbf{M} = \mathcal{C}(\bW_{T^{*}})$. In the original paper, the weights are ranked by their absolute magnitude values, and the bottom $p\%$ are set to zeros in the mask $\mathbf{M}$
    \item \textit{Rewinding.} Rewind the remaining parameters $\mathbf{W}$ to their value in $\mathbf{W}_k$ producing the network $f(\mathbf{x};\mathbf{M}\odot\mathbf{W}_k)$. 
    \item \textit{Iterate.} Until a sparsity or accuracy criterion is met, retrain the resulting sub-network by returning to step 2.(b)
\end{enumerate}

This iterative training method remains costly as it requires to repeatedly train the model several times. In order to address this issue, Morcos et al. \cite{morcos2019one} evaluated the possibility to \textit{transfer} the found tickets across optimizers or datasets. They found that WT indeed appear to learn generic inductive biases which improve training on other datasets.

\subsubsection{Limitations of the lottery ticket approach} \label{sec:lottery_limits}

\begin{figure}[t!]
\begin{center}
\includegraphics[width=.4\textwidth]{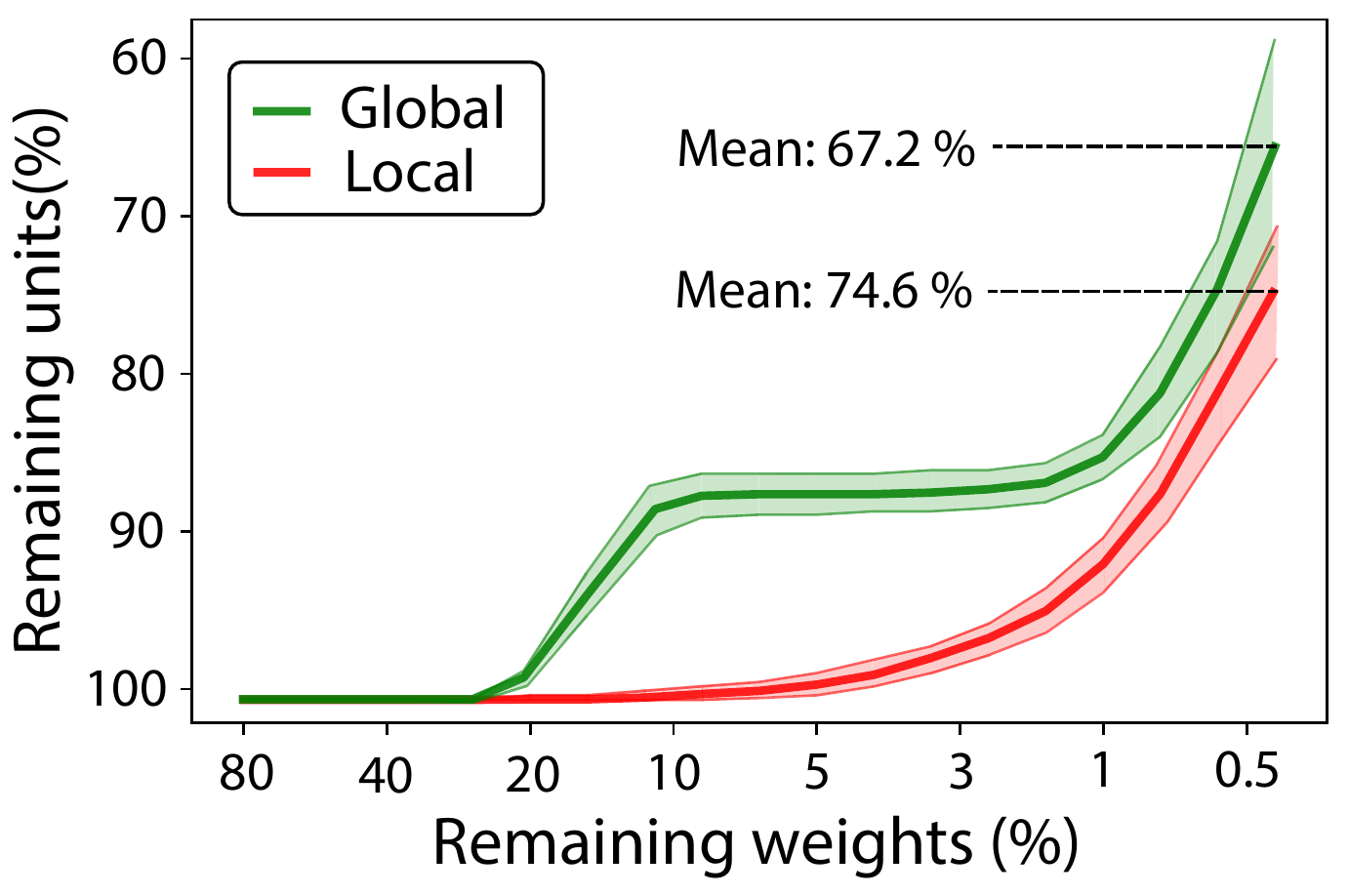}
\caption{Real prunability of a network under \textit{masking} approaches. Even though masking appears to remove up to 99.5\% of the weights, in reality we can only remove a very slight fraction of the units (one third of the network at best in \textit{global} masking)}
\label{fig:pruning_actual}
\end{center}
\end{figure}

\textit{Masking or trimming.} One of the major issues in all unstructured approaches (including the lottery ticket) is that pruning only amounts to \textit{masking} the weight matrix. Hence, this operation hardly produces any network compression, as the model size and inference time remain unchanged. In various papers, the authors propose to obtain true compression by post-processing the uncovered pruning, and remove the units that are entirely masked. In order to estimate the efficiency of this approach, we analyzed this post-processing operation on the original lottery experiment \cite{frankle2018lottery}. We compare the percentage of masked weights and the percentage of units that could truly be pruned, as displayed in Figure~\ref{fig:pruning_actual}. As we can see, there is a huge divergence between the masking ratio (up to 99.5\%), and the real compression (only 25.4\% with local pruning) that is possible with this approach. Hence, the major question we address here is if we could keep the efficiency of masking but perform \textit{real} pruning (termed \textit{trimming} here). Note that a major advantage of trimming is also that each re-training gets iteratively faster, as we effectively remove weights from the network. Hence, the resulting total training time could be largely reduced.

\textit{Local or global.} Another major question in pruning is whether we should rank weights \textit{globally} (across layers) or \textit{locally} (within each layer separately). The advantage of \textit{local} pruning, is that we ensure that all layers preserve an adequate minimal capacity. However, the local criterion cannot modify the relative importance of different layers, and it has been shown that all layers are not equally critical to performance \cite{zhang2019all}. Oppositely, the \textit{global} criterion can freely create bottleneck layers by adapting the size of less important computation. In our case, as we aim to remove entire units, the global pruning reveals an even more critical importance, as it will allow to modify \textit{connectivity} patterns of the network. Indeed, as compared to masking, trimming can not work on connectivity patterns within a layer. Hence, we hypothesize that trimming can only be truly efficient in global setups. Otherwise, this would only amount to performing classical pruning, without truly leveraging the advantages of the lottery ticket hypothesis.

\section{Structured lottery pruning}
Here, we discuss how we can leverage the lottery ticket hypothesis for \textit{structured} pruning. We define criteria that can be used to truly decrease the model size rather than simply masking weights. We introduce a novel criterion based on the mutual information between units and targets. In the following, we use the term \textit{units} to refer to channels or fully-connected units interchangeably.

\subsection{Formalization}

We consider that networks can contain four types of \textit{prunable} layers, namely \textit{linear}, \textit{convolutional}, \textit{recurrent} and \textit{normalization}. We do not detail other types of layers (such as \textit{pooling} and \textit{activation} layers), as they will be unaffected by our trimming strategy. We consider that each layer performs a function $\by = f(\bx; \bW)$ parameterized by a set of weights $\bW$, where the input $\bx\in\R^{N_{in}}$ has dimension $N_{in}$ and the output $\by\in\R^{N_{out}}$ has dimension $N_{out}$. In the case of \textit{trimming}, we need a criterion that returns a sorted list of $N_{out}$ indices, to decide which units to remove. In the following, we will consider both \textit{weight-based} (\textit{magnitude}, \textit{gradient} and \textit{batchnorm}) and \textit{output-based} (\textit{activation} and \textit{information}) criteria. In the case of \textit{output-based} criteria, the list is computed based on the output of each layer. Regarding \textit{weight-based} criteria, we need to adapt the computation for each type of layer.

In the case of linear layers, the operation $f(\bx,\bW)=\bW\bx+\bbb$ implies a simple matrix $\bW\in\R^{N_{out}\times N_{in}}$. Hence, we will compute weight statistics across $j\in[1,N_{in}]$ to obtain $N_{out}$ values. In the convolutional case, the weights are distributed as kernels $\bW_{j}\in\R^{k^{d}}$, with a kernel of size $k$ for $d$-dimensional convolutions. Hence, we will compute statistics over each kernel with $j\in[1,k^{d}]$. Finally, the normalization layers preserve the dimensionality of their inputs with $N_{in}=N_{out}$, and contain a scaling vector $\mathbf{\gamma}\in\R^{1\times N_{in}}$. Apart in the case of the \textit{normalization} criterion, we propagate the trimming criterion to the normalization layers from the layer directly preceding them.

\subsection{Pruning criteria}
\label{sec:pruning_criteria}

We introduce the pruning criteria that are used to rank the units, which can be used for \textit{masking}, but are devised for \textit{trimming}.

\textit{Magnitude.} 
We define a \textit{magnitude-based} criterion, similar to that of the original paper \cite{frankle2018lottery}. However, in our case, we evaluate the overall magnitude of the weights for a complete unit
\begin{equation}
    \mathcal{C}(\bW)= \sum_{j=1}^{N_{in}} \left| W_{i, j} \right|
\end{equation}

\textit{Gradient.} 
In order to see how each weight contribute to the overall solution, we analyze their gradients with respect to the error loss. To do so, we perform a cumulative backward pass on the entire validation dataset to obtain the gradient of the error given each weight and then compute the global unit gradient as
\begin{equation}
    \mathcal{C}(\bW) = \sum_{j=1}^{N_{in}} \left| \frac{\delta \mathcal{L}(\mathcal{D}_{v})}{\delta W_{i, j}} \right|
\end{equation}
with $\mathcal{L}(\mathcal{D}_{v})$ the loss function used for training the network computed on the whole validation dataset $\mathcal{D}_{v}$.

\textit{Activation.}
We can rely on the activation statistics of each unit to analyze their importance. Hence, akin to the previous criterion, we perform a cumulative forward pass through the network after training the model and compute
\begin{equation}
    \mathcal{C}(\bW)=\text{argmin}_{i} \sum_{k=1}^{\mathcal{D}_{v}} \left| f(\bx_{k}, \bW)_{i} \right|
\end{equation}
where we sum across the examples in the validation dataset $\mathcal{D}_{v}$.

\textit{Normalization.}
In this criteria, we rely on the \textit{scaling} coefficient of the normalization layers, as a proxy to determine the importance of the units in the preceding layer $\mathcal{C}(\bW)= \left| \mathbf{\gamma}^{n}_{i} \right|$

\subsection{Mutual information criteria}

Given two random variables $\bx$ and $\by$, with marginal distributions $p(\bx)$ and $p(\by)$ and a joint distribution $p(\bx, \by)$, the \textit{mutual information} (MI) provides a quantitative measure of the \textit{degree of dependency} between these variables. 
\begin{equation}
    I(\bx;\by) = \KL{p(\bx,\by)}{p(\bx)p(\by)},
\end{equation}
where $\KL{p}{q}$ denotes the Kullback-Leibler divergence between distributions $p$ and $q$. Hence, MI measures the divergence between the full joint probability $p(\bx,\by)$ and its factorized version. Therefore, the MI is positive $I(\bx;\by) \geq 0$ and null if $\bx$ and $\by$ are independent variables ($p(\bx,\by)=p(\bx)p(\by)$). 
In our case, MI can inform us on how the representation of each layer $\bz=f(\bx,\bW)$ contains information on the target $\by$, or is independent from it, defining the criterion
\begin{equation}
    \mathcal{C}(\bW)=\text{max}_{i} I(\bz_{i};\by)
\end{equation}
where we compute the output of each unit $\bz_{i}$ on the validation set and compute their MI with the training target $\by$. This criterion is motivated by the fact that we would like to keep units that contain the most information on the target. Unfortunately, MI can only be computed if we have access to the analytic formulation of the distributions. Furthermore, in the case of deterministic networks with continuous variables $\by$ and $\bz$, the MI value $I(\bz;\by)$ is actually infinite. To remedy this problem, the most straightforward approach is to add noise to the hidden activity $\bz'=\bz+\xi$, where $\xi\sim \mathcal{N}(0, \sigma^{2})$ to obtain a finite MI evaluation. In order to approximate the MI, we rely on the Ensemble Dependency Graph Estimator (EDGE) method \cite{noshad2019scalable}, which combines hashing and dependency graphs to provide a non-parametric estimation of MI. 


\subsection{Scaling}

In order to perform global pruning, we need to adequately scale criteria values across layers, to ensure a fair pruning. Indeed, there is no clear bound to the weights or activation values (if we use non-saturating functions). Hence, we explore two scaling operations. First, we perform \textit{normalization} of the criteria values by the maximal value in a given layer. Second, we perform scaling based on the layer dimensionality. This has the advantage of ensuring that we do not remove too much weights in a given layer, while being related to successful initialization schemes, which rely on the \textit{fan in} and \textit{fan out} of each layer.

\section{Experiments}

\subsection{Models}
\label{subsec:models}

In order to evaluate model trimming for generative audio, we reimplemented several state-of-art models. Because of space constraints, we provide minimal details but rely on all implementation choices from the original papers, unless stated otherwise


\textit{Wavenet} introduced by van Oord and al. \cite{van2016wavenet} is implemented as a stack of dilated convolutions to model the raw audio signal as a product of conditional probabilities. We use 2 successive stacks of 20 layers of convolutions with 256 channels and 128 residual channels trained with a cross-entropy loss. The training process relies on teacher forcing, leading to faster training (while the generation remains sequential and slow).

\textit{SING.} proposed by Défossez and al. in \cite{defossez2018sing} is a convolutional neural audio synthesizer that generates waveform given desired categorical inputs. In this paper, we choose to focus on an auto-encoding version of the model used at first to train the decoder, composed of 9 layers of 1D convolution layers with 4096 channels and stride of 256. The encoder mirrors the decoder with similar settings. The architecture is calibrated for 4 second input signals.

\textit{DDSP.}
The Differentiable Digital Signal Processing model has been introduced by Engel and al. in \cite{engel2020ddsp}, and is a lightweight recurrent based architecture for raw audio generation. It implements a reverbered harmonic plus noise synthesizer whose parameters are infered by a gated recurrent unit, based on an input pitch and loudness. We rely on a hidden size of 512 with 3 fully-connected layers and latent size 128 for the Gated Recurrent Units (GRU). The synthesis part uses a filter of size 160 and 100 partials.

\subsection{Datasets}
\label{subsec:datasets}
The models are evaluated by training on the following datasets.

\textit{NSynth} contains 305,979 single notes samples from a range of acoustic and electronic instruments divided into ten categories, as detailed in \cite{engel2017nsynth}. This leads to 1006 instruments, with different pitches at various velocities available as raw waveforms. All notes last 4 seconds with a sampling-rate of 16kHz. As this incurs an extremely large training time, we rely on subsampling, randomly picking 10060 samples (ten notes per instrument). 

\textit{Studio-On-Line} provides individual note recordings sampled at 44100 Hz for 12 orchestral instruments, as detailed in \cite{esling2018generative}. For each instrument, the full tessitura is played for 10 different extended techniques, amounting to around 15000 samples.





For both datasets, we compute the STFT with 5 window sizes ranging from 32 to 1024. Log-magnitudes are computed with a floor value $\epsilon=5e^{-3}$. All datasets are randomly split between \textit{train} (80\%), \textit{valid} (10\%) and \textit{test} (10\%) sets before each training.

\subsection{Training}\label{subsec:section_training}
All models are trained following their respective procedure detailed in each corresponding paper. Hence, hyperparameters vary depending on the models and datasets, but we use a common minibatch size of 64, the ADAM optimizer, a weight decay penalty of $2e^{-4}$, initial learning rate of $1e^{-3}$, which is halved every $10$ non-decreasing epochs. We train each model to completion and perform masking or trimming for $30\%$ of the weights at each iteration. We repeat this process $15$ times, leading to models with up to $99.5\%$ of their original weights removed. 



\section{Results}\label{sec:section_results}
We detail different aspects of our proposal to obtain very light models for generative audio. First, we compare our \textit{trimming} proposal to the original lottery \textit{masking} (Section~\ref{subsec:masking_trimming}), and confirm our hypothesis on the importance of a \textit{global} selection in trimming (Section~\ref{subsec:local_global}). Then, we evaluate the success of the different proposed criteria (Section~\ref{subsec:selection_criteria}) for the \textit{trimming} method with \textit{global} selection. To assess qualitative results, we provide audio samples on the supporting webpage of this paper.



\begin{figure*}[t!]
\begin{center}
\includegraphics[width=.82\textwidth]{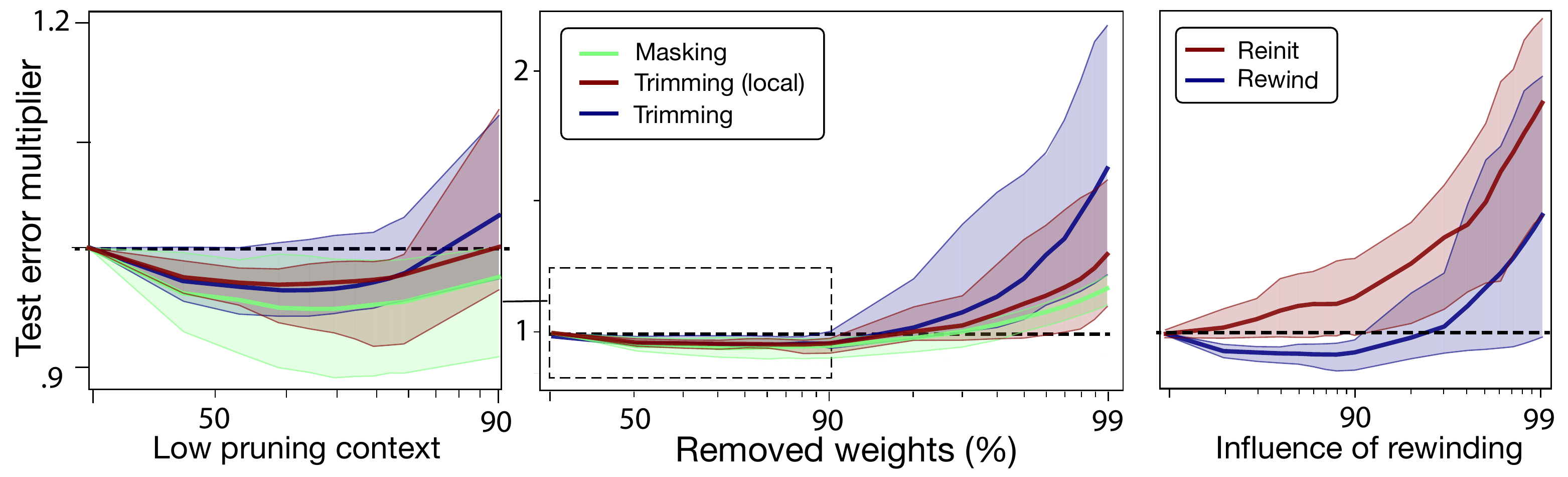}
\caption{Comparison of \textit{masking} and \textit{trimming} in terms of test set error, when iteratively removing weights. We zoom in the error curves (\textit{left}) at low pruning ratios and show the difference between \textit{reinitializing} or \textit{rewinding} the weights (\textit{right}).}
\label{fig:pruning_masking}
\end{center}
\end{figure*}

\subsection{Masking or trimming}
\label{subsec:masking_trimming}

In this section, we evaluate the lottery ticket hypothesis for generative audio and compare the efficiency of pruning based either on \textit{masking} or \textit{trimming} strategies. For \textit{masking}, we use the same setup as the original lottery ticket paper, by using a \textit{magnitude} criteria with a \textit{local} selection \cite{frankle2018lottery}. For \textit{trimming}, we rely on our proposed \textit{information} criterion, using a \textit{global} selection. As a reference point, we also compute the results of \textit{trimming} with a \textit{magnitude} criterion and \textit{local} selection. To observe the effect of model pruning, we compute the mean test error rates across different models as we increasingly prune their weights, using the different selection criteria. As different models rely on different losses and evaluations, we normalize the results by the accuracy obtained by the largest model, and analyze the variation to this reference point. This leads to the \textit{test error multiplier}, which explains the \textit{relative ratio} to the errors across models, regardless of their absolute values. As discussed in Section~\ref{sec:lottery_limits}, there is a huge discrepancy in the effective gain that can be obtained from the masking approach. Nevertheless, we display the results comparing the amount of masking to the amount of trimming, as we seek to maintain the accuracy of the lottery tickets with commensurate amount of pruning. We display this analysis in Figure~\ref{fig:pruning_masking}.

First, as we can see in this figure, we confirm that lottery tickets can be found in generative audio tasks, as shown by the results of the \textit{masking} method. Indeed, in low pruning scenarios, we found smaller models that have a higher accuracy than their larger counterparts, with the lowest mean test error multiplier of \textbf{0.893} being obtained when masking \textbf{76.1\%} of the weights. The error of models remain lower, even if we mask up to \textbf{95\%} of the weights. Then, the error increases, but remains in an acceptable range from the original model, even with up to \textbf{99\%} of the weights masked. When observing the results for the \textit{trimming} method, we can see that we are able to maintain similar results. However, this method leads to a true reduction of the model size and inference time. In order to evaluate more precise aspects of the results, we also closely analyze \textit{low pruning contexts} (up to 90\% of the weights removed). With that comparison, we see that our approach performs in the same range as the original lottery, by providing smaller error rates for low pruning and reaching a factor of \textbf{0.912} when removing \textbf{80\%} of the weights. As an increasing amount of units are removed, the trend seems to be reversed but the trimmed models manage to remain in commensurate accuracy, with a factor of \textbf{1.2} even when removing up to \textbf{99\%} of the weights.  Hence, one of the strongest result in this paper, is that we can obtain models that are more accurate, while being \textasciitilde10 times smaller. An other strong result is that we can reduce models by \textasciitilde100, and still keep a low error rate. Note that the behavior of \textit{global} and \textit{local} depend on \textit{low} or \textit{high} contexts, which we analyze in the next section.

\subsection{Local or global selection in trimming}\label{subsec:local_global}

\begin{figure}[t!]
\begin{center}
\includegraphics[width=.45\textwidth]{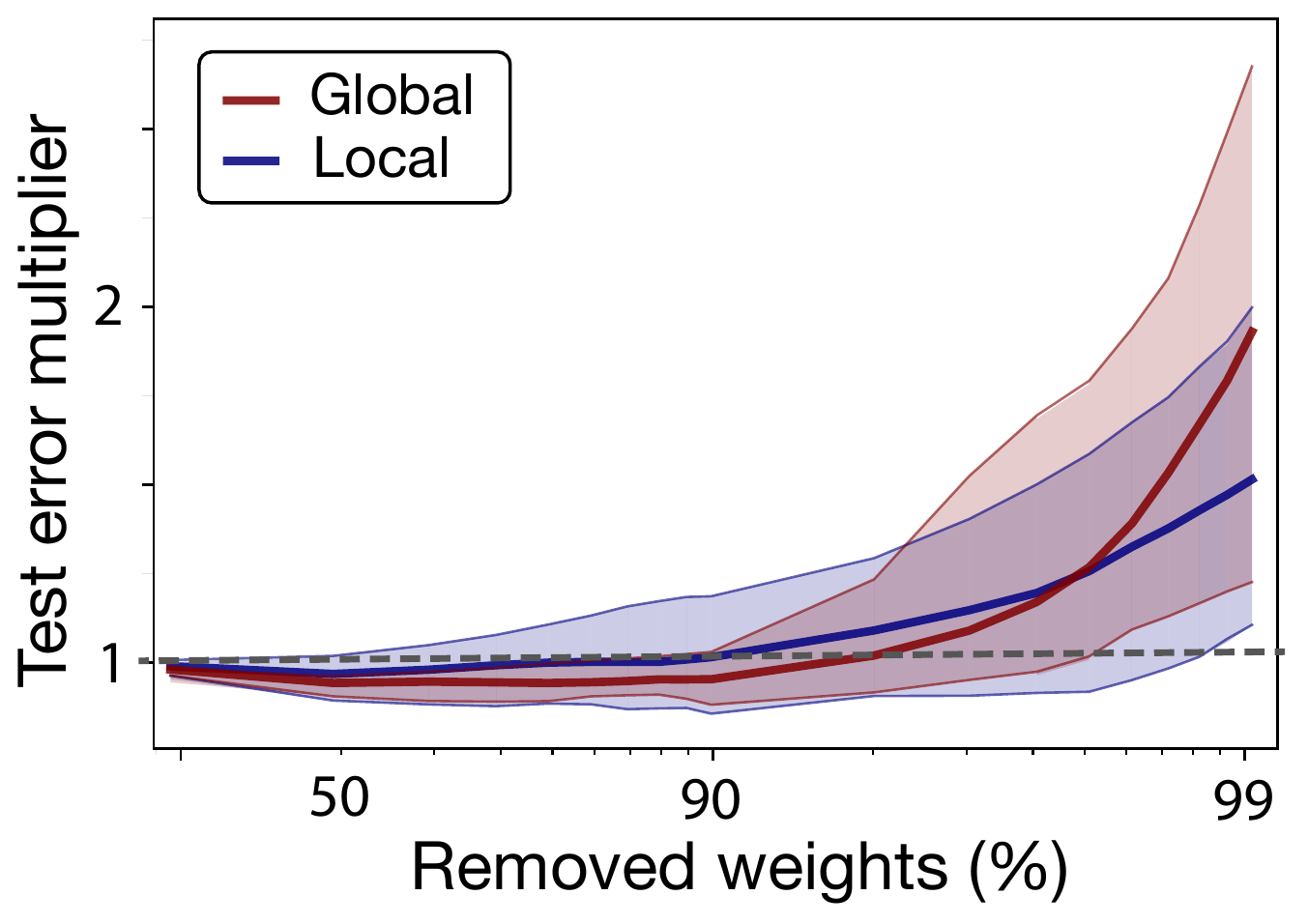}
\caption{Comparison of test error rates between \textit{local} and \textit{global} selection in \textit{trimming}, when iteratively removing weights.}
\label{fig:pruning_local}
\end{center}
\end{figure}

In this section, we evaluate our original hypothesis, that global selection is required to perform efficient \textit{trimming}, whereas local selection can only provide a sub-efficient form of architecture search. Hence, we perform the same analysis as in the previous section, for our \textit{trimming} method across all selection criteria, either for \textit{local} or \textit{global} selection. We display the results of this analysis in Figure~\ref{fig:pruning_local}

As we can see, both selection criteria can provide lower error rates when evaluated at low pruning ratios. This seems to confirm the first part of the lottery ticket hypothesis, even in situations where we effectively remove (trim) units from the network. It appears that the \textit{global} criterion provides lower error rates for \textit{lower} pruning ratios (up to 90\%). This seems to corroborate our initial hypothesis on the crucial importance of using a global selection when performing trimming, to adapt the underlying connectivity. However, as we increase the amount of pruning, the \textit{global} selection quickly degrades, while \textit{local} selection seems to maintain error range. This might come from the fact that \textit{global} selection create bottlenecks too quickly, which causes the performance to degrade. This warrants the need to define more subtle normalization operators, or using global selection in the first phase of compression, to then rely on local for higher pruning contexts.  

\begin{figure*}[t!]
\begin{center}
\includegraphics[width=.9\textwidth]{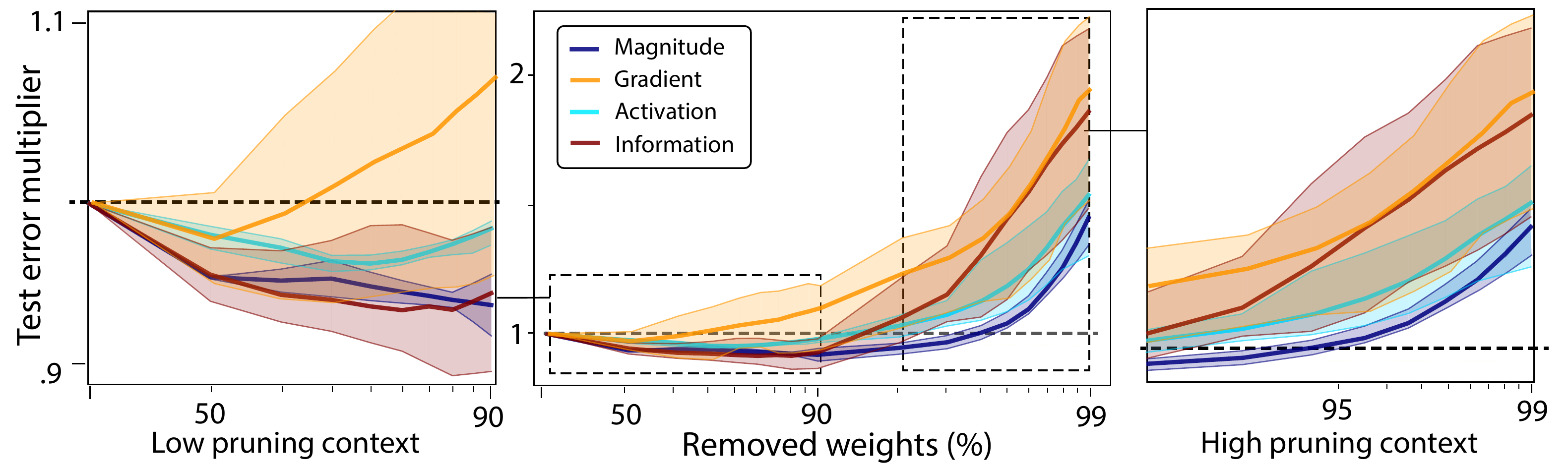}
\caption{Comparison of various pruning criteria in the case of \textit{trimming} with \textit{global} selection in terms of test set error, when iteratively removing weights. We detail two specific parts of the error curves (\textit{left}) at low pruning ratios, we obtain a \textit{lower error} than larger models and (\textit{right}) in high pruning contexts, we obtain extremely small models that still maintain a commensurate error rate.}
\label{fig:pruning_criteria}
\end{center}
\end{figure*}

\subsection{Selection criteria comparison}\label{subsec:selection_criteria}

In this section, we evaluate the efficiency of the various selection criteria proposed in Section~\ref{sec:pruning_criteria}. To do so, we evaluate the full training with the \textit{trimming} method and \textit{global} selection across different models. The results are displayed in Figure~\ref{fig:pruning_criteria}.

As we can see, most criteria can perform an adequate trimming in low pruning contexts. However, when dealing with high pruning scenarios, the differences are more pronounced. In low-pruning situations, our proposed \textit{mutual information} criterion appears to outperform the other ones. With this criterion, the best performing models appear after removing 80\% of the weights and outperform the accuracy of larger models. We are able to remove up to 95\% of the weights without loosing any accuracy, which leads to models that are 20 times smaller, with equivalent quality. However, passed this point it seems that the \textit{information} criterion quickly degrades, whereas other criteria maintain a constant error augmentation. This could be explained by the fact that we are relying on an approximation of the true MI, which can lead to biased estimations. This bias is further worsened by the fact that the evaluation is only performed on subsets of the dimensions and limited number of examples because of memory constraints.

\begin{table}
\begin{centering}
\begin{tabular}{ccccc}
\hline 
\textbf{Model} & \textbf{CPU} & \textbf{FLOPS$^{*}$} & \textbf{Drive} & \textbf{RAM}\tabularnewline
\hline 
\hline 
\multicolumn{2}{l}{\textbf{Arduino}} & \tabularnewline
\hline 
ATMega1280 & 16 MHz & 160 K. & 128 K. & 8 K.\tabularnewline
ATMega2560 & 32 MHz & 320 K. & 256 K. & 16 K.\tabularnewline
\hline 
\hline 
\multicolumn{2}{l}{\textbf{Raspberry Pi}} & \tabularnewline
\hline 
RPi 1B & 700 MHz & 41 M. & 256 M. & 512 M.\tabularnewline
RPi 2B & 900 MHz & 53 M. & 1 G. & 1 G.\tabularnewline
\hline 
\end{tabular}
\par\end{centering}
\caption{Properties of different \textit{Arduino} micro-controllers and \textit{Raspberry Pi} embedded platform (*: FLOPS are inferred values).}
\label{tab:arduino_properties}
\end{table}

\begin{figure*}[t!]
\begin{center}
\includegraphics[width=.95\textwidth]{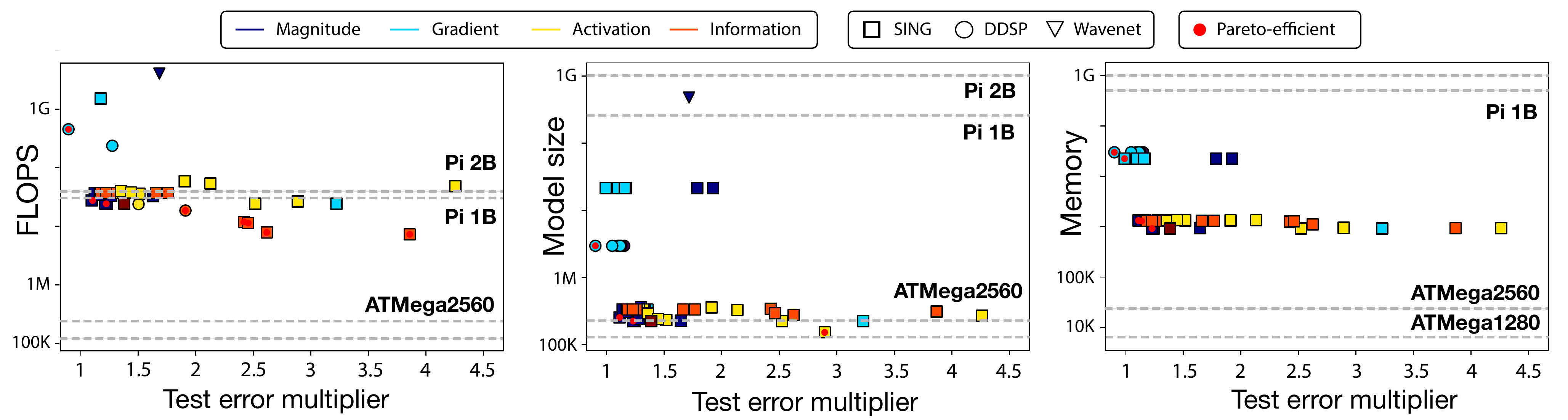}
\caption{Evaluating theoretical embeddability of the light models for deep generative audio on \textit{Arduino} and \textit{Raspberry Pi} platforms.}
\label{fig:embedding_bounds}
\end{center}
\end{figure*}

\section{Embedding deep generative audio}

As discussed earlier, the goal of our aggressive pruning approaches is that we could obtain deep audio models that fit on embedded hardware. However, these systems have very strong constraints, as summarized in Table~\ref{tab:arduino_properties}\footnote{These properties were gathered from the user manuals and the FLOPS are inferred from the listed CPU properties}. 

\subsection{Evaluating theorical embeddability}
In order to assess the performances of light models on embedded architectures, we evaluate aspects specifically pertaining to model compression and complexity with three different metrics.


\textit{Floating point operations (FLOPs)} describes the number of operations required to generate an audio sample at inference time.

\textit{Model disk size} exhibits the size taken by the model (more precisely by its parameters dictionnary) when saved to disk.

\textit{Read-write memory} computes the combined number of memory accesses (read and write operations) when generating a sample.

This measures can be broadly grouped as evaluating either a theoretical \textit{real-time bound} or an \textit{embeddable bound}. The \textit{real-time bound} assess if the model can sample audio in real-time on a given platform. Hence, this is directly measured by the FLOPS required by a single pass of the model to produce one second of audio. The \textit{embeddable bound} measures if the model fit the size requirements of the target platform, both being able to store the model on disk, and having a fitted read-write memory usage.
To evaluate these various constraints, we rely on models that are trimmed using our approach, at their smallest pruning step. We plot these results for every models depending on their requirements and corresponding error rates in Figure~\ref{fig:embedding_bounds}.

Globally speaking, it seems that the models are still quite far from being embeddable on highly constrained hardware such as Arduino. Notably, the memory and FLOPS usage remains largely higher than what the platform can handle. Although it seems that the model size requirements could be obtained for some models, this would come at the price of a highly increased error rates (with the smallest embeddable error being ~2.5 times the original one). We denote in the figure the models that strike an optimal balance (in the Pareto sense) between these two objectives. Several of our lightweight models could be already embedded and run on less constrained hardware, such as Raspberry Pi. Notably, the model size and memory requirements seem to largely fit the constraints, even for the Rasperry Pi 1B. The only issue would relate to FLOPS that seem to be borderline to the CPU capacity. However, more aggressively pruned models could provide a sufficient reduction, with only marginal increases of the error rates.



\section{Conclusions}

In this paper, we devised a method to produce extremely small deep neural networks for generative audio, by leveraging the \textit{lottery ticket} hypothesis. We have shown that this approach could be applied in that context, but that it did not provide gains in the effective size and efficiency of the resulting models. To alleviate these limitations, we developed novel methods of \textit{structured pruning} that allow to truly remove units from the models. We showed that it is possible only by leveraging \textit{global} selection criteria, to adapt the connectivity patterns in the network. This also confirmed the surprising result that smaller models tend to provide higher accuracy than their heavy counterpart. We showed that we could remove up to 95\% of the network without significant loss in accuracy.
Finally, we discussed the possibility of embedding these light models in constrained architectures such as \textit{Arduino} and \textit{Raspberry Pi}, by testing the final model properties against the requirements of the architectures.

\bibliographystyle{IEEEbib}
\bibliography{main} 

\begin{thebibliography}{10}

\bibitem{mehri2016samplernn}
Soroush Mehri, Kundan Kumar, Ishaan Gulrajani, Rithesh Kumar, Shubham Jain,
  Jose Sotelo, Aaron Courville, and Yoshua Bengio,
\newblock ``Samplernn: An unconditional end-to-end neural audio generation
  model,''
\newblock in {\em International Conference on Learning Representations}, 2017.

\bibitem{van2016wavenet}
A{\"a}ron van~den Oord, Sander Dieleman, Heiga Zen, Karen Simonyan, Oriol
  Vinyals, Alex Graves, Nal Kalchbrenner, Andrew Senior, and Koray Kavukcuoglu,
\newblock ``Wavenet: A generative model for raw audio,''
\newblock in {\em 9th ISCA Speech Synthesis Workshop}, 2016, pp. 125--125.

\bibitem{belkin2019reconciling}
Mikhail Belkin, Daniel Hsu, Siyuan Ma, and Soumik Mandal,
\newblock ``Reconciling modern machine-learning practice and the classical
  bias--variance trade-off,''
\newblock {\em Proceedings of the National Academy of Sciences}, vol. 116, no.
  32, 2019.

\bibitem{arora2018optimization}
Sanjeev Arora, N~Cohen, and Elad Hazan,
\newblock ``On the optimization of deep networks: Implicit acceleration by
  overparameterization,''
\newblock in {\em 35th International Conference on Machine Learning}, 2018.

\bibitem{lecun1990optimal}
Yann LeCun, John~S Denker, and Sara~A Solla,
\newblock ``Optimal brain damage,''
\newblock in {\em Advances in neural information processing systems}, 1990, pp.
  598--605.

\bibitem{liu2018rethinking}
Zhuang Liu, Mingjie Sun, Tinghui Zhou, Gao Huang, and Trevor Darrell,
\newblock ``Rethinking the value of network pruning,''
\newblock {\em arXiv preprint arXiv:1810.05270}, 2018.

\bibitem{frankle2019lottery}
Jonathan Frankle, Gintare~Karolina Dziugaite, Daniel~M Roy, and Michael Carbin,
\newblock ``Stabilizing the lottery ticket hypothesis,''
\newblock {\em arXiv preprint arXiv:1903.01611}, 2019.

\bibitem{frankle2020the}
Jonathan Frankle, David~J. Schwab, and Ari~S. Morcos,
\newblock ``The early phase of neural network training,''
\newblock in {\em International Conference on Learning Representations}, 2020.

\bibitem{defossez2018sing}
Alexandre D{\'e}fossez, Neil Zeghidour, Nicolas Usunier, L{\'e}on Bottou, and
  Francis Bach,
\newblock ``Sing: Symbol-to-instrument neural generator,''
\newblock in {\em Advances in Neural Information Processing Systems}, 2018, pp.
  9041--9051.

\bibitem{engel2020ddsp}
Jesse Engel, Lamtharn Hantrakul, Chenjie Gu, and Adam Roberts,
\newblock ``Ddsp: Differentiable digital signal processing,''
\newblock {\em International Conference on Learning Representations}, 2020.

\bibitem{esling2018generative}
Philippe Esling, Axel Chemla-Romeu-Santos, and Adrien Bitton,
\newblock ``Generative timbre spaces with variational audio synthesis,''
\newblock in {\em Proceedings of the International Conference on Digital Audio
  Effects (DAFx)}, 2018.

\bibitem{engel2019gansynth}
Jesse Engel, Kumar~Krishna Agrawal, Shuo Chen, Ishaan Gulrajani, Chris Donahue,
  and Adam Roberts,
\newblock ``Gansynth: Adversarial neural audio synthesis,''
\newblock {\em arXiv preprint arXiv:1902.08710}, 2019.

\bibitem{perraudin2013fast}
Nathana{\"e}l Perraudin, Peter Balazs, and Peter~L S{\o}ndergaard,
\newblock ``A fast griffin-lim algorithm,''
\newblock in {\em IEEE Workshop of Signal Processing to Audio and Acoustics
  (WASPAA)}. IEEE, 2013.

\bibitem{engel2017nsynth}
Jesse Engel, Cinjon Resnick, Adam Roberts, Sander Dieleman, Douglas Eck, Karen
  Simonyan, and Mohammad Norouzi,
\newblock ``Neural audio synthesis of musical notes with wavenet
  autoencoders,''
\newblock {\em International Conference on Machine Learning}, vol. 70, pp.
  1068--1077, 2017.

\bibitem{wang2019neural}
Xin Wang, Shinji Takaki, and Junichi Yamagishi,
\newblock ``Neural source-filter waveform models for statistical parametric
  speech synthesis,''
\newblock {\em IEEE Transactions on Audio, Speech, and Language Processing},
  vol. 28, pp. 402--415, 2019.

\bibitem{li2016pruning}
Hao Li, Asim Kadav, Igor Durdanovic, Hanan Samet, and Hans~Peter Graf,
\newblock ``Pruning filters for efficient convnets,''
\newblock {\em arXiv preprint arXiv:1608.08710}, 2016.

\bibitem{han2015learning}
Song Han, Jeff Pool, John Tran, and William Dally,
\newblock ``Learning both weights and connections for efficient neural
  network,''
\newblock in {\em Advances in neural information processing systems}, 2015, pp.
  1135--1143.

\bibitem{lee2018snip}
Namhoon Lee, Thalaiyasingam Ajanthan, and Philip~HS Torr,
\newblock ``Snip: Single-shot network pruning based on connection
  sensitivity,''
\newblock {\em arXiv preprint arXiv:1810.02340}, 2018.

\bibitem{frankle2018lottery}
Jonathan Frankle and Michael Carbin,
\newblock ``The lottery ticket hypothesis: Finding sparse, trainable neural
  networks,''
\newblock in {\em International Conference on Learning Representations}, 2019.

\bibitem{morcos2019one}
Ari Morcos, Haonan Yu, Michela Paganini, and Yuandong Tian,
\newblock ``One ticket to win them all: generalizing lottery ticket
  initializations across datasets and optimizers,''
\newblock in {\em Advances in Neural Information Processing Systems}, 2019.

\bibitem{zhang2019all}
Chiyuan Zhang, Samy Bengio, and Yoram Singer,
\newblock ``Are all layers created equal?,''
\newblock {\em arXiv:1902.01996 preprint}, 2019.

\bibitem{noshad2019scalable}
Morteza Noshad, Yu~Zeng, and Alfred~O Hero,
\newblock ``Scalable mutual information estimation using dependence graphs,''
\newblock in {\em International Conference on Acoustics, Speech and Signal
  Processing (ICASSP)}. IEEE, 2019, pp. 2962--2966.

\end{thebibliography}

\end{document}